\documentclass{article}
\usepackage{ijcai24}

\usepackage{times}
\usepackage{soul}
\usepackage{url}
\usepackage{pdflscape}
\usepackage[hidelinks]{hyperref}
\usepackage[utf8]{inputenc}
\usepackage[small]{caption}
\usepackage{graphicx}
\usepackage{amsmath}
\usepackage{amsthm}
\usepackage{booktabs}
\usepackage{multirow}
\usepackage{colortbl}
\usepackage{threeparttable}
\usepackage{algorithm}
\usepackage{fp}
\usepackage[table]{xcolor}
\usepackage{collcell}
\usepackage{algorithmic}
\usepackage[switch]{lineno}
\usepackage{amssymb}
\usepackage{amsmath}
\usepackage{tcolorbox}
\usepackage{lipsum}

\linenumbers

\title{\textit{Disce aut Deficere}: Evaluating LLMs Proficiency on the INVALSI Italian Benchmark}

\author{
Fabio Mercorio$^{1,3}$\and
Mario Mezzanzanica$^{1,3}$\and
Daniele Potertì$^{2}$\and
Antonio Serino$^{2}$\And
Andrea Seveso$^{1,3}$\\
\affiliations
$^1$Dept of Statistics and Quantitative Methods, University of Milano Bicocca, Italy\\
$^2$Dept of Economics, Management and Statistics, University of Milano Bicocca, Italy\\
$^3$CRISP Research Centre \url{crispresearch.eu},  University of Milano Bicocca, Italy\\
}

\begin{document}
\nolinenumbers

\maketitle

\begin{abstract}
    Recent advancements in Large Language Models (LLMs) have significantly enhanced their ability to generate and manipulate human language, highlighting their potential across various applications. Evaluating LLMs in languages other than English is crucial for ensuring their linguistic versatility, cultural relevance, and applicability in diverse global contexts, thus broadening their usability and effectiveness. We tackle this challenge by introducing a structured benchmark using the INVALSI tests, a set of well-established assessments designed to measure educational competencies across Italy. 
    Our study makes three primary contributions: Firstly, we adapt the INVALSI benchmark for automated LLM evaluation, which involves rigorous adaptation of the test format to suit automated processing while retaining the essence of the original tests. Secondly, we provide a detailed assessment of current LLMs, offering a crucial reference point for the academic community. Finally, we visually compare the performance of these models against human results.
    Additionally, researchers are invited to submit their models for ongoing evaluation\footnote{\url{https://huggingface.co/spaces/Crisp-Unimib/INVALSIbenchmark}}, ensuring the benchmark remains a current and valuable resource.
\end{abstract}

\section{Introduction}
\label{sec:introduction}

In recent years, Large Language Models (LLMs) have emerged as a pivotal advancement in the field of Natural Language Processing (NLP) and Artificial Intelligence (AI)~\cite{chang_survey_2023}. 
Their remarkable capacity to generate human-like text, comprehend context, and perform various language-related tasks, such as question-answering and summarisation, has led to transformative applications across industries. 
Model evaluation is paramount but difficult since there are various important qualities to consider: models should be precise, resilient, fair, and efficient, among others~\cite{liang2022holistic}.

Developing language models that function effectively across diverse global languages and evaluating them remains a significant and ongoing challenge~\cite{srivastava2022beyond}. The currently available models often perform highly in English but are lacking in underrepresented languages~\cite{ruder2021xtreme}. This is due to factors such as the scarce and lower quality available data~\cite{kreutzer2022quality}, smaller contributing communities, and Anglo-centric cultural bias in development~\cite{talat2022you}. 

The INVALSI (National Institute for the Evaluation of the Education and Training System) test has been essential to Italy's educational assessment framework since the 2005-2006 academic year. It aims to evaluate students' competencies across various subjects, including Italian language and mathematics, at multiple stages of the education system in a standardised manner. The test's primary objective is to assess students' linguistic proficiency, a crucial skill that schools are tasked with developing as per national educational guidelines~\cite{trinchero2014servizio}.

The Italian section of the test measures various aspects of language mastery, such as reading comprehension (understanding, interpreting, and evaluating written text), grammatical knowledge (recognising and applying grammatical structures correctly) and lexical competence (using and understanding vocabulary appropriately within context).

The INVALSI test is structured to evaluate these linguistic domains through a series of tasks that reflect real-world language use. This involves understanding texts of varying complexities~\cite{corsini2013validita}, utilising vocabulary appropriately~\cite{toth2023riflettere}, and applying grammatical rules to construct and analyse sentences~\cite{guzzo2023competenza}. The test design considers the progressive complexity appropriate for each educational level, ensuring that the assessment is both challenging and fair, corresponding to the expected developmental stage of the students.
Additionally, it aims to provide a transparent benchmark for student performance in Italian, offering detailed feedback that can guide further instructional strategies~\cite{pastore2017questione}. This feedback is crucial for teachers and educational authorities to identify areas where students excel or need more support, enhancing overall educational outcomes.

Since the test covers a wide range of linguistic and comprehension skills~\cite{corsini2013rilevazioni}, using it to evaluate LLMs can provide a detailed view of a model's proficiency in handling real-world, nuanced language tasks designed for human learners. The test's structured and standardised nature makes it an excellent benchmark for comparing different LLMs with questions culturally and contextually relevant to Italian speakers. However, our findings are relevant across all languages since they assess various general capabilities, such as word formation and text comprehension abilities. Also, since it is designed for different educational stages, it provides varied levels of complexity and challenge. This aspect can gauge an LLM's capability at different difficulty levels, reflecting its potential scalability and adaptability across simpler to more complex linguistic tasks.

Given these robust evaluation criteria, this paper aims to establish a benchmark for assessing large language models by leveraging the INVALSI framework.  

\subsection{Contributions}

The contributions of this work are three-fold:

\begin{enumerate}
    \item We structure the INVALSI test, a notable national test for Italian students, as an automated evaluation benchmark for LLMs\footnote{We use a subset of tests, handpicked from different years and educational levels, ensuring to exclude those with questions that are difficult to rephrase or that require analysing images.};
    \item We conduct a thorough analysis of existing models, establishing a reference for the research community;
    \item We visually display results across several important metrics and compare models' performances to human standards, pinpointing the strengths and weaknesses.
\end{enumerate}

The remainder of the paper is structured as follows: Section 2 presents related work in state of the art; Section 3 details our data curation process for creating the benchmark; Section 4 displays the results of multiple models tested against this benchmark. Section 5 discusses these results and identifies limitations; Section 6 concludes the paper and outlines proposals for future work.

\section{Related Work}
\label{sec:context}

A Large Language Model is a deep learning model trained on vast amounts of text data to develop a sophisticated understanding of language structures and semantics. Leveraging the transformer architecture~\cite{vaswani_attention_2017}, LLMs employ self-attention mechanisms to process sequential data efficiently. The emergence of pretraining and fine-tuning strategies, exemplified by models like BERT~\cite{devlin_bert_2019} and GPT~\cite{radford_improving_2018}, has enabled the development of LLMs with impressive capabilities. These models are trained on massive text corpora, allowing them to learn intricate linguistic patterns and generate coherent and contextually relevant text. 

\paragraph{Multilingual models.}
LLMs have also shown multilingual capabilities based on their training on multilingual data~\cite{touvron2023llama2,li2024eliciting} and vocabulary~\cite{pires2019multilingual,chung2020improving,liang2023xlm}. 
GPT-3 and its successors have shown different capabilities in several languages~\cite{armengol2021multilingual} since their training corpora are, in part, composed of non-English texts.
For example, Bloom~\cite{le2023bloom} was trained in 46 different natural languages, showing emerging capabilities in multilingual tasks.
Most recently, also smaller size models~\cite{jiang2024mixtral,touvron2023llama2}, due to the inclusion of multilingual data in the training process, have shown emerging capabilities in German, French, Spanish and Italian, but not performing as well in the most prominent training language.

\paragraph{Italian models.}
Despite the emergence of many multilingual models, their Italian language capabilities lack consistency. The particularity of the Italian language and the need for models with higher capabilities in Italian tasks have started a challenge to create Italian-specific language models. 
The pioneer of this challenge is Geppetto~\cite{de2020geppetto}, which proposes the first Italian-adapted language model based on GPT-2 small architecture. 
The emergence of several methodologies to adapt large models with low resources~\cite{hu2021lora,dettmers2024qlora} has allowed the adapting of larger models to the Italian language.
Recent attempts to adapt LLMs to the Italian Language use LLaMa as a base model:~\cite{santilli2023camoscio} instruct-tune LLaMa using the Alpaca dataset translated to Italian. In contrast,~\cite{bacciu2023fauno} propose a Parameters Efficient Fine Tuning~\cite{peft} of the same base model, using a synthetically generated and machine-translated dataset.~\cite{basile2023llamantino} performed a PEFT on LLaMa-2, proposing models of sizes 7B, 13B and 70B. Most recently ~\cite{polignano2024advanced} propose a LLaMa 8B PEFT adaptation to Italian.

\paragraph{Available benchmarks.}
The evaluation of LLMs is performed to assess various capabilities.  
For commonsense reasoning, ~\cite{clark2018think} releases a dataset composed of English language science exam questions drawn from a variety of sources, ~\cite{zellers2019hellaswag} comprises multiple choice questions about grounded situations, with four answer choices about what might happen next in the scene. In contrast, ~\cite{sakaguchi2021winogrande} is a collection of problems formulated as a fill-in-a-blank task with binary options, where the goal is to choose the right option for a given sentence, which requires commonsense reasoning.
To assess multi-step mathematical reasoning capabilities ~\cite{cobbe2021training} published a benchmark of high-quality linguistically diverse grade school math word problems. ~\cite{lin2021truthfulqa} is a Question-Answering benchmark to measure whether a language model is truthful in generating answers to questions. The benchmark comprises questions that span 38 categories, including health, law, finance and politics. To evaluate multitask accuracy ~\cite{hendrycks2020measuring} propose a benchmark that covers 57 tasks in different domains, including elementary mathematics, US history, computer science, law, etc. Finally~\cite{rajpurkar2016squad} aims to assess the model's reading comprehension capabilities, proposing a benchmark consisting of questions posed by crowd workers on a set of Wikipedia articles, where the answer to every question is a segment of text, or span, from the corresponding reading passage, or the question might be unanswerable. 

\paragraph{Italian benchmarks.}
The Italian NLP community lacks the depth of original language evaluation benchmarks compared to the English community. Some previously cited benchmarks, such as~\cite{hendrycks2020measuring,zellers2019hellaswag,clark2018think}, are commonly used to evaluate LLMs in Italian after being automatically translated.
Benchmarks natively Italian are less common.
\cite{basile2023uinauil} propose a Unified Benchmark for Italian Natural Language Understanding that covers textual entailment, Event detection and classification, factuality classification, sentiment polarity classification, irony detection and hate speech detection. \cite{lai2023evalita} proposes a collaborative benchmark on 13 tasks. Both benchmarks focus on classification-based tasks and do not explore LLM properties like common-sense reasoning. Another Italian benchmark is represented by~\cite{landro2022two}, covering only Italian news text summarisation abilities. These benchmarks lack a wide range of possible scenarios to evaluate LLMs, thus not allowing a comprehensive evaluation~\cite{liang2022holistic}.

\section{INVALSI Benchmark Curation}
\label{sec:methods}
\subsection{Data Collection}
We have collected from public sources 58 unique tests, divided into 141 unique units, with 2114 questions and 2808 unique items. Some questions are subdivided into multiple items, each requiring a specific answer. 

The data has been gathered from the Gestinv\footnote{\url{https://www.gestinv.it/Index.aspx}} database~\cite{bolondi2017database}.
The Gestinv database is available to teachers, schools, students, and families. It includes thousands of questions from national standardised assessments and collects materials from 2008. These materials include test booklets, overall and individual question results, statistical analyses, reports, and other documentation. The Gestinv database is used in research on mathematics education and professional development programs for teachers, both in-service and training. Their goal is to provide tools to effectively utilise the vast amount of information that the National Assessments provide on the learning achievements of Italian students. The database is widely used in professional development programs for teachers in Italian schools and various initial university training courses.

Questions' formatting is sometimes not adequately structured for LLM evaluation; for instance, it is sometimes impossible to automatically transcribe the questions into structured fields, necessitating further inspection of images and PDFs. For this reason, we also collected corresponding PDF files and images. Manual inspection was required to ensure accuracy. In cases where questions involved graphical elements, we modified them into an appropriate multiple-choice format. For example, if the task required a student to find and underscore a word, we reformulated the question to allow selection from multiple choices. Similarly, if the task involved drawing a line between two groups of concepts—a common task for younger students—we rephrased it to involve choosing the correct association from given options. Generally, we aimed to adapt the questions to a format that allows the model to select the correct answer from a pool of choices if it aligns with the original type of question. Tab.~\ref{tab:qualitative_examples} shows a few original and formatted question examples.


\definecolor{lightblue}{rgb}{0.68, 0.85, 0.90}
\definecolor{lightgreen}{rgb}{0.68, 0.85, 0.45}
\definecolor{lightred}{rgb}{0.96, 0.68, 0.68}
\begin{table*}
\centering
\footnotesize
\begin{tabular}{|p{5cm}|p{5cm}|p{5cm}|}
\hline
\vspace{0.1pt}
\textbf{Original MC Question} \vspace{0.1pt}& \vspace{0.1pt}\textbf{Original MCC Question} \vspace{0.1pt}& \vspace{0.1pt}\textbf{Original RU Question} \vspace{0.1pt}
\\
\parbox{5cm}{In the sentence "Livia was running in the park when a strong storm broke out", what are the events indicated by the two verbs?
\begin{itemize}
    \item They are contemporary and have the same duration
    \item They are contemporary and indicate habitual actions
    \item The first event occurs during the second event
    \item The second event occurs during the first event
\end{itemize}} &
\parbox{5cm}{Read this sentence: “The night bird made such an acute sound that frightened the inhabitants of the forest very much".\\
    Indicate in the table which words are nouns and which are not.\\ Put a cross for each row of the table.\\
\begin{itemize}
    \item The -- It's a name / It's not a name
    \item night -- It's a name / It's not a name
    \item ...
\end{itemize}}
 &
\parbox{5cm}{Where would you put the letter h? If necessary, write it in the square.\\
\\
\(\square\)avevo perso l’autobus così arrivai tardi \(\square\)a scuola.} \\
\hline
\vspace{0.1pt}
\textbf{Prompted MC Question} \vspace{0.1pt}& \vspace{0.1pt}\textbf{Prompted MCC Question} \vspace{0.1pt}& \vspace{0.1pt}\textbf{Prompted RU Question} \vspace{0.1pt} \\
\parbox{5cm}{\colorbox{yellow}{Question:} \\
In the sentence "Livia was running in the park when a strong storm broke out", what are the events indicated by the two verbs? \\

\colorbox{yellow}{Options:} 

['A. They are contemporary and have the same duration', \\'B. They are contemporary and indicate habitual actions', \\'C. The first event occurs during the second event', \\'D. The second event occurs during the first event'] \\

\colorbox{yellow}{Instructions:} \\
\fcolorbox{yellow}{yellow}{\parbox{\dimexpr\linewidth-2\fboxsep-2\fboxrule}{You must return the letter corresponding to the correct answer in square brackets.}} \\
\colorbox{yellow}{Response format: [letter]} \\
\colorbox{yellow}{Answer:} \vspace{0.1pt}}
 & \parbox{5cm}{\colorbox{yellow}{Question:} \\
Read this sentence: “The night bird made such an acute sound that frightened the inhabitants of the forest very much".\\

Indicate whether this word is a noun or not:\\
                The \\
                
\colorbox{yellow}{Options:} 

['A. It's a name',\\ 'B. It's not a name'] \\

\colorbox{yellow}{Instructions:} \\
\fcolorbox{yellow}{yellow}{\parbox{\dimexpr\linewidth-2\fboxsep-2\fboxrule}{You must return the letter corresponding to the correct answer in square brackets.}} \\
\colorbox{yellow}{Response format: [letter]} \\
\colorbox{yellow}{Answer:} \vspace{0.1pt}} & 
\parbox{5cm}{\colorbox{yellow}{Question:} \\
 Where would you put the letter h? Indicates whether or not it is necessary instead of **.\\

**avevo perso l’autobus così arrivai tardi a scuola. \\

\colorbox{yellow}{Options:} 

['A. Necessary',\\ 'B. Not necessary'] \\

\colorbox{yellow}{Instructions:} \\
\fcolorbox{yellow}{yellow}{
\parbox{\dimexpr\linewidth-2\fboxsep-2\fboxrule}{You must return the letter corresponding to the correct answer in square brackets.}} \\
\colorbox{yellow}{Response format: [letter]} \\
\colorbox{yellow}{Answer:} \vspace{0.1pt}} \\
\hline
\vspace{0.1pt}
\textbf{Model 1 Response}\vspace{0.1pt} & \vspace{0.1pt}
\textbf{Model 1 Response}\vspace{0.1pt} & \vspace{0.1pt}
\textbf{Model 1 Response}\vspace{0.1pt} \\

\colorbox{lightgreen}{[D]} \vspace{0.1pt} &
\colorbox{lightgreen}{[B. It's not a name]} \vspace{0.1pt}&
\colorbox{lightgreen}{[B]} \vspace{0.1pt}\\

\hline
\vspace{0.1pt}
\textbf{Model 2 Response}\vspace{0.1pt} & \vspace{0.1pt}
\textbf{Model 2 Response}\vspace{0.1pt} & \vspace{0.1pt}
\textbf{Model 2 Response}\vspace{0.1pt} \\
\fcolorbox{lightred}{lightred}{\parbox{\dimexpr\linewidth-2\fboxsep-2\fboxrule}{[C. The first event occurs during the second event]}}\vspace{0.1pt} &
\colorbox{lightgreen}{B. It's not a name}\vspace{0.1pt} & \colorbox{lightred}{[A. Necessary]}\vspace{0.1pt} \\
\hline
\end{tabular}
\caption{Examples of different types of questions translated from the original Italian form, along with the prompted format, correct and incorrect model answers. 
The RU question was not translated as it pertains to a morphological task unique to the Italian language. In the benchmark, all questions are originally composed in Italian. The translation is provided solely for the convenience of the reader.}
\label{tab:qualitative_examples}
\end{table*}

\subsection{Dataset Characteristics}

We have selected 10 tests comprising 31 unique units, 409 questions, and 625 items from the above data. A test comprises two or more different units; each question can have more than one item to answer. The sample of tests has been chosen by manual inspection, aiming to include different grades and years and avoiding those with questions that require inspecting an image or contained questions that would be hard to reformulate for language model comprehension.

\begin{table}[tbh]
    \centering
    \footnotesize
    \begin{tabular}{lrrr}
        \toprule
        School Grade & \# Tests & \# Questions & \# Items \\
        \midrule
        2nd Grade (Primary School) & 2 & 34 & 72\\
        5th Grade (Primary School) & 2 & 75 & 117\\
        6th Grade (Middle School) & 1 & 87 & 118\\
        8th Grade (Middle School) & 2 & 86 & 88\\
        10th Grade (High School) & 2 & 77 & 139 \\
        13th Grade (High School) & 1 & 50 & 91 \\
        \bottomrule
    \end{tabular}
    \caption{Distribution of tests, questions and items by educational grade.}
    \label{tab:test_selection}
\end{table}

Each question in the benchmark is labelled with the specific lexical macro area it aims to assess.
Tab.~\ref{tab:question_type_aspetto_distribution} shows each macro area distribution in our benchmark.

\begin{table}[tbh]
\centering
\footnotesize
\begin{tabular}{lp{3.5cm}r}
\toprule
Section & Macro Area & \# Questions\\
\midrule
\multirow{3}{*}{\parbox{1.3cm}{Text comprehension}}&Reconstruct the meaning of the text, locally or globally & 179 (43.8\%) \\
&Locate and identify information within the text  & 108 (26.4\%) \\
&Reflect on the content or form of the text, locally or globally, and evaluate them & 33 (8.1\%)\\ 
\midrule
\multirow{6}{*}{\parbox{1.3cm}{Reflection on the language}} & Lexicon and semantics & 29 (7.1\%)\\
& Morphology & 24 (5.9\%) \\
& Syntax  & 19 (4.6\%)  \\
& Word formation  & 7 (1.7\%) \\
& Textuality and pragmatics  & 6 (1.5\%) \\
& Spelling  & 4 (1.0\%)\\
\bottomrule
\end{tabular}
\caption{Distribution of questions by section and macro area.}
\label{tab:question_type_aspetto_distribution}
\end{table}

In particular, the questions labelled with the first three elements in Tab. \ref{tab:question_type_aspetto_distribution} are related to text comprehension skills. \textit{"Locate and identify information within the text"} is used for all the questions aiming to evaluate the capability of identifying several kinds of information within the provided context. \textit{"Reconstruct the meaning of the text, locally or globally"} label is related to questions that aim to assess the ability to reconstruct, starting from the text, the context in which it is inserted and the reader's 'encyclopaedic' knowledge, the set of meanings that the text conveys, together with how they are conveyed: in other words, the logical-conceptual and formal organisation of the text itself, in any case concerning the context. Lastly, questions labelled as \textit{"Reflect on the content or form of the text, locally or globally, and evaluate them"} aim to evaluate the ability to interpret texts and their shape, expressing an evaluation. 
The remaining macro areas are designed and structured to evaluate grammatical knowledge. More specifically,  \textit{"Word formation"} aims to evaluate knowledge about base words and their derivatives and \textit{"Lexicon and semantics"} aims to assess knowledge about the semantic relationship between words. The questions belonging to \textit{"Morphology"} category aim to check the competencies with several lexical categories (noun, adjective, verb, etc.) and sub-categories (possessive adjective, proper name, etc.). In contrast, the use of accents and apostrophes, upper and lower case letters, word segmentation, use of doubles, and cases of mismatch between phonemes and graphemes (use of h, q, digrams, etc.) is evaluated by the questions categorised as \textit{Spelling}. All the questions within \textit{Syntax} aim to evaluate the correctness of syntactic rules of Italian written language, and \textit{"Textuality and pragmatics"} aims to assess signs of text organisation and cohesion phenomena.

The questions come in five distinct formats, which are:

\begin{table}[tbh]
\centering
\begin{tabular}{lrr}
\toprule
Format & \# Questions & \# Items \\
\midrule
MC  & 334 (81.7\%) & 340  \\
MCC & 38 (9.3\%) & 228 \\
RU  & 33 (8.1\%) & 50 \\
CL  & 3 (0.7\%) & 6 \\
RB  & 1 (0.2\%) & 1 \\
\bottomrule
\end{tabular}
\caption{Distribution of questions by format.}
\label{tab:question_type_distribution}
\end{table}

\begin{itemize}
    \item \textit{Multiple Choice (MC}): composed of a question with several answer options, among which only one is correct. It is the most common question format in the selected tests, comprising 334 questions (81.7\% of the total) and 340 items. The variation in numbers arises because some questions demand the completion of two distinct choices, both of which must be correct. The answer choices are typically four—labelled A, B, C, and D.
    \item \textit{Multiple Complex Choice (MCC)}: composed of input questions and multiple items to answer. It is the second most common type of question, with 38 (9.3\%) instances and 228 items. One answer from among the 2 or more available options must be given for each item, and only one is correct. The question is deemed correct only if all the items are rightly answered.
    \item \textit{Unique Response (RU)}: involves open-ended questions in which there are no options or suggestions and where only one answer is considered correct (with sometimes a limited number of possible variants). We found 33 (8.1\%) RU questions and 50 items in the selected tests.
    \item \textit{Short Response (RB)}: is a specific sub-type of RU questions in which the answer is short, tending to consist of a single identifying word. Only one (0.2\%) question and item in our benchmark are classified as RB.
    \item \textit{Cloze (CL)}: can be open-ended or closed-ended. In both cases, a text with missing words is provided. The aim is to complete the text by inserting the correct missing terms. Three (0.7\%) of such questions are available, consisting of two items each. In the case of open answers, no hints and no options are provided, whereas, in the case of closed answers, a set of possible words are provided to be inserted within the text provided.
\end{itemize}

\subsection{Evaluation}

The diversity in question formats ensures a comprehensive evaluation of the models' capabilities. 81.7\% of the questions follow a standard multiple-choice format with closed options, 9.3\% of the questions are multiple binary choices, requiring the model to affirm or deny statements, and 1\% of the questions require specific handling, such as adding punctuation to a specific sentence.
As seen in Tab.~\ref{tab:question_type_distribution}, 90\% of the questions in the benchmark dataset created are MC and MCC. Consequently, evaluation involves verifying whether the generated answer includes the target answer, accomplished through regular expressions. In the case of questions with multiple items, such as MCC, the question is only considered correct if all the items are answered correctly.
We included instructions in the prompt of each question, directing the LLM to format its response within square brackets. Subsequently, we used pattern matching to determine if the answer given was accurate. The benchmark tests models in a zero-shot setting by giving them unseen multiple-choice questions from various subjects. The models must understand the question and provide the correct answer based on their general knowledge or the narrative passage proposed by the question.

Specific strategies for other questions, such as RU, RB, and CL, were developed to cater to their unique requirements. These questions demand more than selecting an available option; thus, the prompts were tailored with additional or specific instructions to ensure correct responses. Each function for generating prompts checks for the presence of context and choices, incorporating them into the prompt if available. This ensures the LLM has all pertinent information to answer the question accurately.

Evaluation functions for assessing model responses utilise various methods tailored to the type of response expected. Regular expressions often parse responses from a designated format, such as content enclosed in square brackets, and compare them directly to the correct answers. These comparisons can include tactics like case insensitivity or omitting certain characters to account for minor formatting variations. Specific methods used in evaluations include:

\begin{itemize}
    \item \textit{Word Matching.} Extracting words or phrases and comparing them directly to the answers. These methods are particularly suited for questions expecting specific terms or phrases as responses.
    
    \item \textit{Pattern Matching.} Employing regular expressions to identify patterns within the model's output, aiding in evaluating syntactic or grammatical responses.

    \item \textit{Semantic Comparison.} We employ BERTScore~\cite{zhang2019bertscore} to evaluate the semantic content of responses with a threshold of 0.7 for correctness. This approach is vital for complex language tasks where paraphrasing or diverse expressions may still be correct. It ensures that the essence of the response aligns with the intended meaning, even if the exact words used differ from the model answers. The threshold was set after a thorough empirical evaluation, manually validating all model answers. Unlike traditional metrics like BLEU~\cite{papineni2002bleu}, which rely on exact n-gram matching and often fail to capture semantic similarity, BERTScore leverages contextualised embeddings from pre-trained language models like BERT. These embeddings capture the nuances of word meaning within the context of the entire sentence, allowing for a more accurate assessment of semantic equivalence between the generated response and the reference. 
\end{itemize}

\section{Results}
\label{sec:results}

\subsection{Model Selection Criteria}
We perform our evaluation using a variety of notable foundational or fine-tuned models. The models considered in this study have been chosen according to the following characteristics:

\begin{itemize}
    \item \textit{Parameter Threshold.} Models with at least three billion parameters are included in the survey. This parameter threshold acts as a filter to encompass LLMs with substantial complexity and capacity for language comprehension and generation.
    \item \textit{Temporal Range.} The selection focuses on models published from 2022 onwards, a period marked by a significant surge in advancements and the establishment of influential models in the field of LLMs.
    \item \textit{Institutional Source.} Models published by prominent organisations such as OpenAI and Meta are considered. This criterion ensures that the models chosen emanate from authoritative entities renowned for their contributions to AI research.
    \item \textit{Popular Italian Models.} For models specifically trained or fine-tuned in the Italian language, the selection process adapts to the dynamic nature of the field.
\end{itemize}

\paragraph{Closed-source models.} We include OpenAI's GPT-3.5 and GPT-4~\cite{achiam2023gpt}, both recognised for their advanced language capabilities. Additionally, we consider Anthropic's Claude series, which includes Haiku, Sonnet, and Opus, each excelling in text generation tasks. Also part of our evaluation is Google’s Gemini Pro 1.0~\cite{team2023gemini} and 1.5~\cite{reid2024gemini}, along with Cohere's Command R+.

\paragraph{Open-source models.}
Our selection of open-source models includes Mistral 7B~\cite{jiang2023mistral} and Mixtral~\cite{jiang2024mixtral}. Furthermore, we examine Meta's LLaMA 3 series, specifically the 8B and 70B models~\cite{llama3modelcard}.

\paragraph{Italian open-source models.}
For models specifically tuned to the Italian language, we include Minerva 3B, a foundational model trained from scratch in Italian and English on over 600 billion tokens. We also consider LLaMAntino 3~\cite{polignano2024advanced}, a model fine-tuned from LLaMa 3, and Zefiro 7B~\cite{tunstall2023zephyr}. We did not inspect older versions of Italian models due to their lower performance.

\paragraph{Model size.} We also categorise the models into three categories by their size. \textbf{Small (S)} models have fewer than 8 billion parameters, or for the closed models accessible via API; they cost less than 0.50\$ per million input tokens. \textbf{Medium (M)} models can go up to 70 billion parameters for open-source versions and cost less than 5\$ per million input tokens for proprietary APIs. \textbf{Large (L)} models exceed these limits.

\subsection{Model Performance}

\begin{figure}[tbh]
    \centering
\includegraphics[width=0.99\columnwidth]{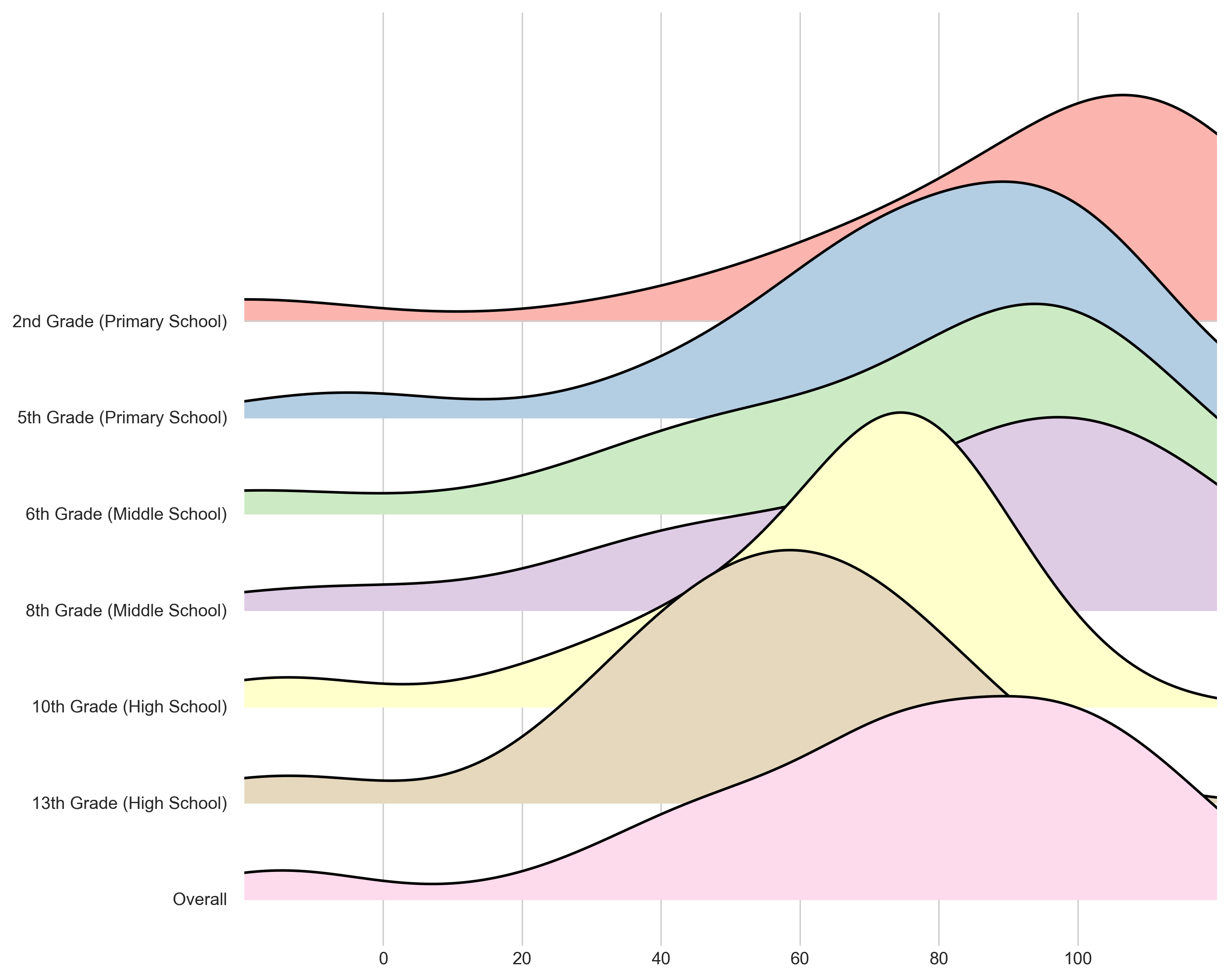}
    \caption{Visualising the accuracy of various models across different school grades. Each layer represents a different grade level, from 2nd grade in primary school to 13th grade in high school, showing the distribution of performance accuracy for each grade.} 
    \label{fig:grade_model_performance}
\end{figure}

Given the various dimensions available, we present an overall accuracy distribution for each model to conduct our evaluation. The school grade is the most critical variable influencing our analysis; in Fig.~\ref{fig:grade_model_performance}, we provide a plot illustrating how accuracy distributions vary across different grades. This visual does not detail specific numbers but instead offers a general sense of how performance shifts with grade level, with a more detailed analysis to follow.

\begin{figure}[tbh]
    \centering
    \includegraphics[width=0.99\columnwidth]{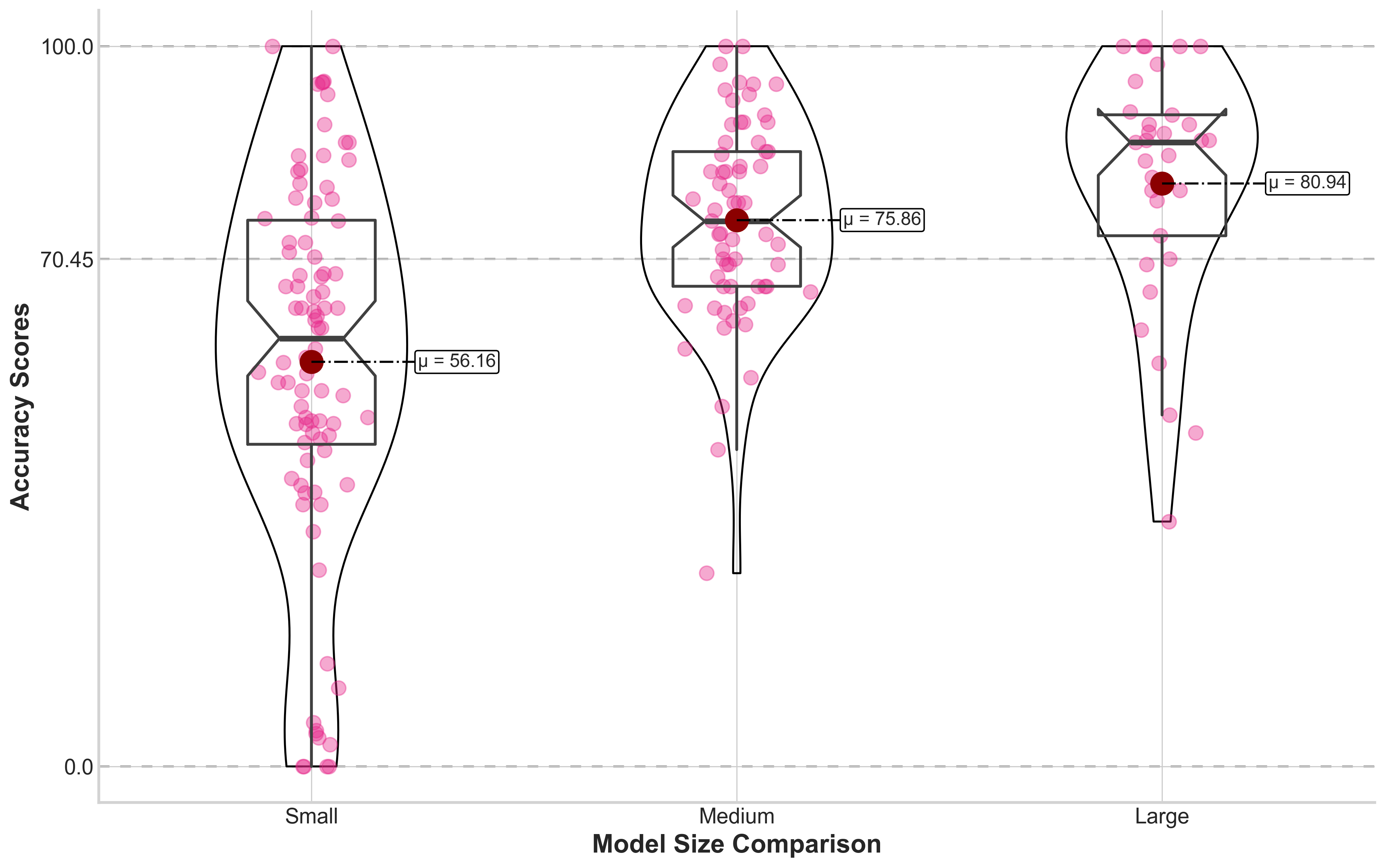}
    \caption{Distribution of accuracy scores of language models categorised by size: small, medium, and large. Each plot represents the distribution of accuracy scores within each category, with individual data points highlighted, each representing a test taken by a model, and the mean accuracy marked by a horizontal line.}
    \label{fig:model_size_comparison}
\end{figure}

\paragraph{Impact of model size.}
Another intriguing dimension to consider is the impact of model size on performance. We now present in Fig.~\ref{fig:model_size_comparison} the distribution of scores segmented by the model size. Recall from the previous subsection how the models are categorised based on size.

\paragraph{Detailed analysis of question format and macro areas.}
We now delve into a detailed analysis of how question format influences performance. In Tab.~\ref{tab:accuracies_1} and~\ref{tab:accuracies_2}, we present the accuracy scores for each model, stratified by both school grade and question format. Due to the stratification and the limited number of questions in some categories, extreme values such as 100 or 0 are more attainable in the sections with few items. The number of questions for each category is indicated in the table headers.
The model average is in the last column of Tab.~~\ref{tab:accuracies_2}. 
Similarly, Tab.~\ref{tab:macro_aspetto} shows the performance comparison of AI models across linguistic macro areas. 

\begin{table*}[htb!]
    \centering
    \scriptsize
    
\def\cca#1{\cellcolor{teal!#1}\ifnum #1>50\color{white}\fi{#1}}

\begin{tabular}{l|rr|rrrrr|rrr}
\toprule
School Grade & \multicolumn{2}{c}{2nd Grade (Primary School)} & \multicolumn{5}{c}{5th Grade (Primary School)} & \multicolumn{3}{c}{6th Grade (Middle School)} \\
\midrule
Question Format (\#) & MC (32) & MCC (2) & CL (1) & MC (60) & MCC (7) & RB (1) & RU (6) & MC (71) & MCC (7) & RU (9) \\
\midrule
claude-3-haiku & \cca{100}.0 & \cca{50}.0 & \cca{0}.0 & \cca{91}.7 & \cca{28}.6 & \cca{0}.0 & \cca{33}.3 & \cca{84}.5 & \cca{57}.1 & \cca{77}.8  \\
claude-3-sonnet & \cca{100}.0 & \cca{100}.0 & \cca{100}.0 & \cca{96}.7 & \cca{85}.7 & \cca{100}.0 & \cca{50}.0 & \cca{88}.7 & \cca{57}.1 & \cca{66}.7  \\
claude-3-opus & \cca{100}.0 & \cca{100}.0 & \cca{100}.0 & \cca{98}.3 & \cca{71}.4 & \cca{100}.0 & \cca{33}.3 & \cca{93}.0 & \cca{85}.7 & \cca{88}.9  \\
command-r-plus & \cca{90}.6 & \cca{0}.0 & \cca{100}.0 & \cca{88}.3 & \cca{14}.3 & \cca{0}.0 & \cca{50}.0 & \cca{80}.3 & \cca{57}.1 & \cca{66}.7  \\
gemini-pro & \cca{96}.9 & \cca{0}.0 & \cca{0}.0 & \cca{90}.0 & \cca{14}.3 & \cca{0}.0 & \cca{16}.7 & \cca{80}.3 & \cca{71}.4 & \cca{66}.7  \\
gemini-flash-1.5 & \cca{90}.6 & \cca{0}.0 & \cca{0}.0 & \cca{86}.7 & \cca{71}.4 & \cca{100}.0 & \cca{33}.3 & \cca{93}.0 & \cca{85}.7 & \cca{88}.9  \\
gemini-pro-1.5 & \cca{96}.9 & \cca{0}.0 & \cca{0}.0 & \cca{90}.0 & \cca{42}.9 & \cca{100}.0 & \cca{33}.3 & \cca{87}.3 & \cca{42}.9 & \cca{77}.8  \\
gpt-3.5-turbo-0125 & \cca{84}.4 & \cca{0}.0 & \cca{0}.0 & \cca{73}.3 & \cca{14}.3 & \cca{0}.0 & \cca{50}.0 & \cca{53}.5 & \cca{42}.9 & \cca{44}.4  \\
gpt-4-turbo & \cca{100}.0 & \cca{100}.0 & \cca{100}.0 & \cca{91}.7 & \cca{71}.4 & \cca{100}.0 & \cca{66}.7 & \cca{63}.4 & \cca{100}.0 & \cca{88}.9  \\
gpt-4o & \cca{78}.1 & \cca{100}.0 & \cca{100}.0 & \cca{83}.3 & \cca{71}.4 & \cca{100}.0 & \cca{66}.7 & \cca{66}.2 & \cca{85}.7 & \cca{77}.8  \\
llama-3-8b-instruct & \cca{65}.6 & \cca{0}.0 & \cca{0}.0 & \cca{66}.7 & \cca{0}.0 & \cca{0}.0 & \cca{16}.7 & \cca{57}.8 & \cca{28}.6 & \cca{11}.1  \\
llama-3-70b-instruct & \cca{96}.9 & \cca{0}.0 & \cca{0}.0 & \cca{90}.0 & \cca{14}.3 & \cca{0}.0 & \cca{33}.3 & \cca{87}.3 & \cca{71}.4 & \cca{66}.7  \\
mistral-7b-instruct & \cca{71}.9 & \cca{0}.0 & \cca{0}.0 & \cca{66}.7 & \cca{0}.0 & \cca{0}.0 & \cca{16}.7 & \cca{59}.1 & \cca{14}.3 & \cca{33}.3  \\
mixtral-8x7b-instruct & \cca{96}.9 & \cca{0}.0 & \cca{0}.0 & \cca{76}.7 & \cca{14}.3 & \cca{0}.0 & \cca{16}.7 & \cca{80}.3 & \cca{57}.1 & \cca{55}.6  \\
LLaMAntino-3-8B & \cca{71}.9 & \cca{0}.0 & \cca{0}.0 & \cca{70}.0 & \cca{14}.3 & \cca{0}.0 & \cca{16}.7 & \cca{67}.6 & \cca{42}.9 & \cca{22}.2  \\
zefiro-7b-base-ITA & \cca{56}.2 & \cca{0}.0 & \cca{0}.0 & \cca{55}.0 & \cca{0}.0 & \cca{0}.0 & \cca{16}.7 & \cca{56}.3 & \cca{0}.0 & \cca{33}.3  \\
Minerva-3B-base-v1.0 & \cca{0}.0 & \cca{0}.0 & \cca{0}.0 & \cca{13}.3 & \cca{0}.0 & \cca{0}.0 & \cca{0}.0 & \cca{0}.0 & \cca{0}.0 & \cca{0}.0  \\
\midrule
\midrule
Models Avg & \cca{82}.2 & \cca{26}.5 & \cca{29}.4 & \cca{78}.1 & \cca{31}.1 & \cca{35}.3 & \cca{32}.4 & \cca{70}.5 & \cca{52}.9 & \cca{56}.9  \\
\bottomrule
\end{tabular}
    \caption{Performance (accuracy \%) comparison of AI models across school grades and question formats for grades 2 to 6.}
    \label{tab:accuracies_1}
\end{table*}

\begin{table*}[htb!]
    \centering
    \scriptsize
    
\def\cca#1{\cellcolor{teal!#1}\ifnum #1>50\color{white}\fi{#1}}

\begin{tabular}{l|rrr|rrrr|rr|r}
\toprule
School Grade & \multicolumn{3}{c}{8th Grade (Middle School)} & \multicolumn{4}{c}{10th Grade (High School)} & \multicolumn{2}{c}{13th Grade (High School)} & All Grades \\
\midrule
Question Format (\#) & MC (81) & MCC (1) & RU (4) & CL (2) & MC (48) & MCC (13) & RU (14) & MC (42) & MCC (8) & Overall \\
\midrule
claude-3-haiku & \cca{85}.2 & \cca{100}.0 & \cca{75}.0 & \cca{50}.0 & \cca{75}.0 & \cca{46}.1 & \cca{64}.3 & \cca{71}.4 & \cca{12}.5 & \cca{78}.0 \\
claude-3-sonnet & \cca{87}.7 & \cca{0}.0 & \cca{75}.0 & \cca{50}.0 & \cca{81}.2 & \cca{53}.9 & \cca{64}.3 & \cca{78}.6 & \cca{12}.5 & \cca{83}.1 \\
claude-3-opus & \cca{93}.8 & \cca{0}.0 & \cca{100}.0 & \cca{50}.0 & \cca{85}.4 & \cca{61}.5 & \cca{71}.4 & \cca{90}.5 & \cca{25}.0 & \cca{88}.5 \\
command-r-plus & \cca{85}.2 & \cca{0}.0 & \cca{100}.0 & \cca{50}.0 & \cca{79}.2 & \cca{46}.1 & \cca{57}.1 & \cca{61}.9 & \cca{12}.5 & \cca{75}.1 \\
gemini-pro & \cca{88}.9 & \cca{0}.0 & \cca{100}.0 & \cca{0}.0 & \cca{79}.2 & \cca{46}.1 & \cca{64}.3 & \cca{69}.0 & \cca{0}.0 & \cca{76}.5 \\
gemini-flash-1.5 & \cca{88}.9 & \cca{0}.0 & \cca{100}.0 & \cca{50}.0 & \cca{81}.2 & \cca{38}.5 & \cca{50}.0 & \cca{81}.0 & \cca{0}.0 & \cca{80}.9 \\
gemini-pro-1.5 & \cca{87}.7 & \cca{0}.0 & \cca{100}.0 & \cca{50}.0 & \cca{79}.2 & \cca{46}.1 & \cca{85}.7 & \cca{85}.7 & \cca{12}.5 & \cca{81}.2 \\
gpt-3.5-turbo-0125 & \cca{67}.9 & \cca{0}.0 & \cca{75}.0 & \cca{50}.0 & \cca{68}.8 & \cca{46}.1 & \cca{71}.4 & \cca{52}.4 & \cca{0}.0 & \cca{61}.1 \\
gpt-4-turbo & \cca{92}.6 & \cca{0}.0 & \cca{100}.0 & \cca{50}.0 & \cca{87}.5 & \cca{61}.5 & \cca{50}.0 & \cca{64}.3 & \cca{12}.5 & \cca{79}.5 \\
gpt-4o & \cca{80}.2 & \cca{0}.0 & \cca{100}.0 & \cca{0}.0 & \cca{68}.8 & \cca{38}.5 & \cca{78}.6 & \cca{38}.1 & \cca{12}.5 & \cca{69}.2 \\
llama-3-8b-instruct & \cca{42}.0 & \cca{0}.0 & \cca{0}.0 & \cca{0}.0 & \cca{54}.2 & \cca{15}.4 & \cca{28}.6 & \cca{57}.1 & \cca{0}.0 & \cca{47}.9 \\
llama-3-70b-instruct & \cca{79}.0 & \cca{0}.0 & \cca{75}.0 & \cca{0}.0 & \cca{68}.8 & \cca{46}.1 & \cca{71}.4 & \cca{76}.2 & \cca{0}.0 & \cca{75}.6 \\
mistral-7b-instruct & \cca{50}.6 & \cca{0}.0 & \cca{25}.0 & \cca{0}.0 & \cca{50}.0 & \cca{23}.1 & \cca{28}.6 & \cca{57}.1 & \cca{0}.0 & \cca{50}.6 \\
mixtral-8x7b-instruct & \cca{71}.6 & \cca{0}.0 & \cca{75}.0 & \cca{0}.0 & \cca{68}.8 & \cca{30}.8 & \cca{57}.1 & \cca{69}.0 & \cca{0}.0 & \cca{68}.5 \\
LLaMAntino-3-8B & \cca{55}.6 & \cca{100}.0 & \cca{50}.0 & \cca{0}.0 & \cca{64}.6 & \cca{23}.1 & \cca{57}.1 & \cca{45}.2 & \cca{0}.0 & \cca{56}.0 \\
zefiro-7b-base-ITA & \cca{43}.2 & \cca{0}.0 & \cca{0}.0 & \cca{0}.0 & \cca{41}.7 & \cca{15}.4 & \cca{42}.9 & \cca{54}.8 & \cca{0}.0 & \cca{44}.3 \\
Minerva-3B-base-v1.0 & \cca{8}.6 & \cca{0}.0 & \cca{0}.0 & \cca{0}.0 & \cca{6}.2 & \cca{0}.0 & \cca{0}.0 & \cca{4}.8 & \cca{0}.0 & \cca{4}.9 \\
\midrule
\midrule
Models Avg & \cca{71}.1 & \cca{11}.8 & \cca{67}.6 & \cca{23}.5 & \cca{67}.0 & \cca{37}.6 & \cca{55}.5 & \cca{62}.2 & \cca{5}.9 & \cca{65}.9  \\
\bottomrule
\end{tabular}
    \caption{Performance (accuracy \%) comparison of AI models (cont.), for grades 8 to 13 and overall average.}
    \label{tab:accuracies_2}
\end{table*}

\begin{table*}[htb!]
    \centering
    \scriptsize
    
\def\cca#1{\cellcolor{purple!#1}\ifnum #1>50\color{white}\fi{#1}}

\begin{tabular}{l|rrr|rrrrrr|r}
\toprule
Section & \multicolumn{3}{c}{Text Comprehension} & \multicolumn{6}{c}{Reflection on the Language} & Both \\
\midrule
Macro Area (\#) & LI (108) & RM (179) & RC (33) & WF (7) & LS (29)  & MO (24) & SP (4) & SY (19) & TP (6) & Overall \\
\midrule
claude-3-haiku & \cca{78}.7 & \cca{86}.0 & \cca{75}.8 & \cca{71}.4 & \cca{65}.5 & \cca{62}.5 & \cca{0}.0 & \cca{57}.9 & \cca{83}.3 & \cca{78}.0 \\
claude-3-sonnet & \cca{87}.0 & \cca{90}.5 & \cca{75}.8 & \cca{100}.0 & \cca{62}.1 & \cca{75}.0 & \cca{0}.0 & \cca{52}.6 & \cca{100}.0 & \cca{83}.1 \\
claude-3-opus & \cca{91}.7 & \cca{91}.6 & \cca{78}.8 & \cca{100}.0 & \cca{82}.8 & \cca{75}.0 & \cca{50}.0 & \cca{89}.5 & \cca{83}.3 & \cca{88}.5 \\
command-r-plus & \cca{74}.1 & \cca{80}.5 & \cca{81}.8 & \cca{71}.4 & \cca{65}.5 & \cca{66}.7 & \cca{0}.0 & \cca{57}.9 & \cca{83}.3 & \cca{75}.1 \\
gemini-pro & \cca{78}.7 & \cca{82}.1 & \cca{81}.8 & \cca{71}.4 & \cca{51}.7 & \cca{70}.8 & \cca{0}.0 & \cca{68}.4 & \cca{66}.7 & \cca{76}.5 \\
gemini-flash-1.5 & \cca{83}.3 & \cca{85}.5 & \cca{81}.8 & \cca{85}.7 & \cca{62}.1 & \cca{83}.3 & \cca{25}.0 & \cca{63}.2 & \cca{66}.7 & \cca{80}.9 \\
gemini-pro-1.5 & \cca{90}.7 & \cca{87}.7 & \cca{84}.8 & \cca{57}.1 & \cca{55}.2 & \cca{58}.3 & \cca{25}.0 & \cca{63}.2 & \cca{33}.3 & \cca{81}.2 \\
gpt-3.5-turbo-0125 & \cca{61}.1 & \cca{64}.8 & \cca{63}.6 & \cca{42}.9 & \cca{55}.2 & \cca{58}.3 & \cca{0}.0 & \cca{47}.4 & \cca{83}.3 & \cca{61}.1 \\
gpt-4-turbo & \cca{77}.8 & \cca{82}.1 & \cca{75}.8 & \cca{71}.4 & \cca{82}.8 & \cca{75}.0 & \cca{50}.0 & \cca{73}.7 & \cca{100}.0 & \cca{79}.5 \\
gpt-4o & \cca{64}.8 & \cca{69}.8 & \cca{51}.5 & \cca{100}.0 & \cca{69}.0 & \cca{87}.5 & \cca{0}.0 & \cca{89}.5 & \cca{100}.0 & \cca{69}.2 \\
llama-3-8b-instruct & \cca{48}.1 & \cca{53}.6 & \cca{63}.6 & \cca{14}.3 & \cca{34}.5 & \cca{29}.2 & \cca{0}.0 & \cca{31}.6 & \cca{50}.0 & \cca{47}.9 \\
llama-3-70b-instruct & \cca{83}.3 & \cca{85}.5 & \cca{75}.8 & \cca{71}.4 & \cca{55}.2 & \cca{33}.3 & \cca{0}.0 & \cca{47}.4 & \cca{50}.0 & \cca{75}.6 \\
mistral-7b-instruct & \cca{51}.9 & \cca{59}.2 & \cca{51}.5 & \cca{28}.6 & \cca{37}.9 & \cca{29}.2 & \cca{0}.0 & \cca{31}.6 & \cca{33}.3 & \cca{50}.6 \\
mixtral-8x7b-instruct & \cca{74}.1 & \cca{77}.1 & \cca{69}.7 & \cca{42}.9 & \cca{37}.9 & \cca{50}.0 & \cca{0}.0 & \cca{52}.6 & \cca{50}.0 & \cca{68}.5 \\
LLaMAntino-3-8B & \cca{60}.2 & \cca{63}.1 & \cca{78}.8 & \cca{28}.6 & \cca{37}.9 & \cca{16}.7 & \cca{0}.0 & \cca{26}.3 & \cca{50}.0 & \cca{56}.0 \\
zefiro-7b-base-ITA & \cca{50}.0 & \cca{49}.7 & \cca{48}.5 & \cca{57}.1 & \cca{20}.7 & \cca{16}.7 & \cca{0}.0 & \cca{26}.3 & \cca{50}.0 & \cca{44}.3 \\
Minerva-3B-base-v1.0 & \cca{4}.6 & \cca{3}.9 & \cca{9}.1 & \cca{28}.6 & \cca{3}.5 & \cca{4}.2 & \cca{0}.0 & \cca{5}.3 & \cca{0}.0 & \cca{4}.9 \\
\midrule
\midrule
Models Avg & \cca{68}.2 & \cca{71}.3 & \cca{67}.6 & \cca{61}.3 & \cca{51}.7 & \cca{52}.5 & \cca{8}.8 & \cca{52}.0 & \cca{63}.7 & \cca{65}.9  \\
\bottomrule
\end{tabular}
    \caption{Performance (accuracy \%) comparison of AI models across macro areas. Categories are abbreviated as: \textit{LI}: Locate and identify information within the text.
    \textit{RM}: Reconstruct the meaning of the text, locally or globally.
    \textit{RC}: Reflect on the content or form of the text, locally or globally, and evaluate them.
    \textit{WF}: Word formation.
    \textit{LS}: Lexicon and semantics.
    \textit{MO}: Morphology.
    \textit{SP}: Spelling.
    \textit{SY}: Syntax.
    \textit{TP}: Textuality and pragmatics.}
    \label{tab:macro_aspetto}
\end{table*}

\subsection{Comparison with Human Respondents}

In evaluating the performance of language models, a critical comparison arises between the responses generated by these models and those of human respondents. We aim to provide insights into the capabilities of language models relative to average human performance.

Not every test we had included the percentage of human accuracies. Specifically, data was available from one test for grade 2; for grades 5 and 6, there were accuracies from two tests each; and for grades 8 and 10, accuracies were available from one test each. Unfortunately, no data on human accuracies was available for grade 13.

\begin{figure*}[btp]
    \centering
    \includegraphics[width=0.9\textwidth]{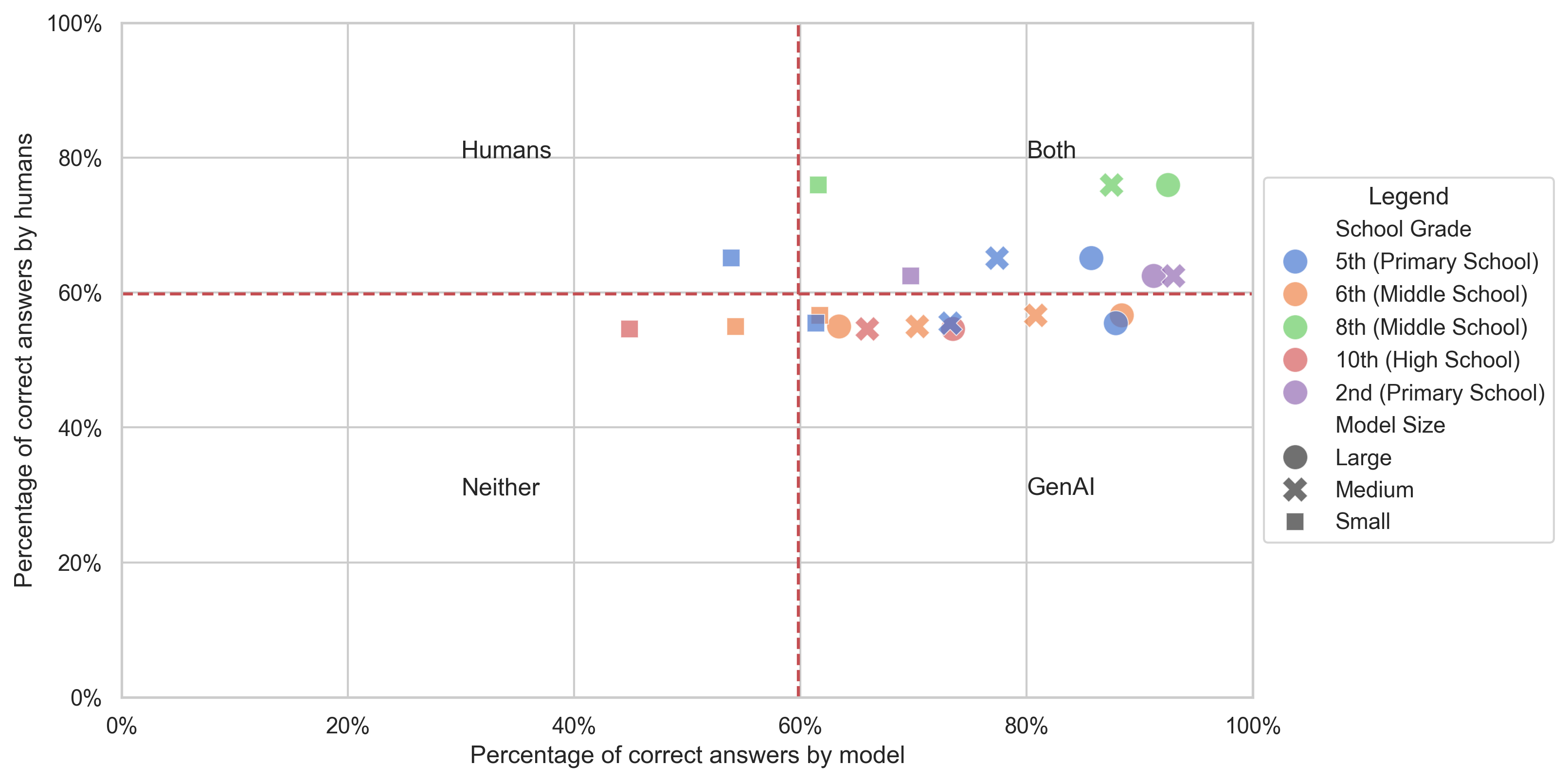}
    \caption{Scatter plot visualising the accuracy of both human respondents and language models on various tests across different grade levels. The red lines represent the median accuracy of human answers at 59.8\%. The graph is divided into four quadrants to categorise the performance: top-right quadrant ("Both"), where both humans and models perform well; top-left quadrant ("Humans"), where humans outperform models; bottom-right quadrant ("GenAI") where models outperform humans; and bottom-left quadrant ("Neither") where neither models nor humans perform well. Each symbol represents the average performance for each model size on a test, and colour coding corresponds to the educational grade level, providing an overview of where AI competes or lags behind human performance. Multiple data points with the same colour and symbol are shown wherever multiple tests for the same school grade exist.}
    \label{fig:quadrants}
\end{figure*}

In Fig.~\ref{fig:quadrants}, we compare human and model performances. The red lines represent the median of human answers, set at 59.8, to delineate which classes of models perform above this benchmark. This division creates four quadrants: both perform well, neither perform well, humans perform better, and models perform better.

\section{Discussion}
\label{sec:discussion}

\paragraph{Model performance across stratifications.}
In analysing the performance results of these models, it is evident that models with a higher number of parameters generally demonstrate superior performance compared to those with fewer parameters, as illustrated in Figure \ref{fig:model_size_comparison}. The figure reveals smaller models exhibit greater variance and dispersion in their accuracy scores than medium and large models. The notches in the violin plots indicate that the confidence intervals for the small and medium models do not overlap. The medium models have a higher average performance with average accuracy scores of 56.16\% for small models and 75.86\% for medium models. Tthe notches for the large models also do not overlap with those of the small and medium models, and they achieve the highest average accuracy score of 80.94\%. This observation is consistent with the scaling laws of language models~\cite{kaplan2020scaling}.


Examination of  Tables~\ref{tab:accuracies_1} and~\ref{tab:accuracies_2} reveals significant variability in model performance across various school grades, especially in MC questions, which are the most prevalent type. The models tend to perform better in lower grades while showing decreased accuracy in higher ones, indicating that the difficulty of the tasks or the complexity of the language used in tests may significantly influence model performance.

A comparative analysis of LLMs across various macro areas is shown in Tab.~\ref{tab:macro_aspetto}. A notable strength of these models is their ability to reconstruct the meaning of text, locally and globally (RM), with an overall accuracy of 71.3\%. Conversely, none of the models consistently performed well in the spelling (SP) category, with an overall accuracy of 8.8\%. Overall, LLMs exhibit superior performance on average in Text Comprehension tasks compared to Reflection on the Language tasks. This aligns with previous findings in the field of Language Understanding, where these language models can excel at understanding context and drawing inferences based on large contexts because of their generative pre-training and discriminative fine-tuning~\cite{radford_improving_2018}. Conversely, syntax and morphology tasks require precise, rule-based understanding and application. Although models can produce grammatically correct text, they often encounter difficulties with tasks that demand explicit knowledge of linguistic rules and higher levels of reasoning, areas where large language models (LLMs) are known to have limitations \cite{valmeekam2022large,huang2022towards}.
None of the models tested could correctly answer all, or even the majority, of the SP category questions. These questions presented a classical Italian writing task:

\begin{tcolorbox}[colback=white!95!black, colframe=white!75!black, title=Spelling Question]
\textbf{Where would you place the letter \textit{h}?} \\ 
If needed, write it in the box.\\

\(\square\)avevo perso l’autobus così arrivai tardi \(\square\)a scuola.
\end{tcolorbox}

The letter "h" should be placed correctly to answer this question to form the appropriate Italian words. The letter "h" is essential in Italian for distinguishing between certain homophones. The correct placement would differentiate "ho" (I have) and "ha" (he/she has), but in this context, the correct form is "a scuola" (at school), meaning no "h" is needed. Therefore, the sentence reads: "avevo perso l’autobus così arrivai tardi a scuola" (I missed the bus so I arrived late at school).
We observed that some of the larger closed models (Claude-3-Opus, Gemini-Flash-1.5, Gemini-Pro-1.5, GPT-4-Turbo) correctly answered one or two of the four spelling questions.

\paragraph{Impact of model size on performance.}
Large closed-source models achieve superior accuracy on benchmarks, successfully addressing approximately 80-85\% of the tasks. The Claude class model exhibits the highest accuracy with Opus at 88.5\%, while the Gemini class models perform optimally with Pro-1.5, achieving an accuracy of 80.9\%. Among OpenAI's GPT class, the 4-Turbo model shows the most proficient performance with an accuracy of 79.5\%\footnote{The performance of GPT class models is influenced by content flagging, as discussed in Section~\ref{sec:censorship}.}.

In the domain of open weights models, the performance typically ranges from 65-75\% across various tasks. Notably, the LLaMA-3-70b-instruct model achieves a 75.6\% success rate, followed closely by Command-R+ at 75.1\%, and mixtral 8x7b at 68.5\%.
Smaller models, possessing 7-5 billion parameters, generally perform around 50\%, with mistral-7b-instruct reaching 50.6\% and LLaMA-3-8b-instruct at 47.9\%. Among the Italian finetuned models, the LLaMAntino-3-8B, which is a finetuned variant of LLaMA-3-8b-instruct, demonstrates a performance improvement, achieving a 56\% success rate—this represents an 8.1 percentage point increase over its base model. Additionally, the zefiro-7b-base-ITA model records a performance of 44.3\%.

Overall, the inference for this benchmark with closed-weight models incurred a cost slightly below \$50. The dataset comprises approximately 620,000 input million tokens, yielding an average output of 17,000 tokens per model. As of the inference date, the costliest model was claude-3-opus, priced at \$15 per million input tokens and \$75 per million output tokens.

\paragraph{Comparison of models with human respondents.}
In Figure~\ref{fig:quadrants}, we compare the overall accuracy of humans with models for each school grade. The x-axis represents the performance of different models, categorised by size with the symbol and school grade with the colour, while the y-axis indicates the performance of humans.

Since the red lines illustrate the average student performance across all grades and tests, half are expected to score above and half below this average by design. Nevertheless, comparing the models' performance is intriguing, particularly about school grades and model size. As anticipated and consistent with previous graphs, models tend to perform worse at higher school grades, while larger models generally show better performance. Interestingly, there is no correlation between school grade and human performance; the highest human performance is recorded in the 8th grade.

\paragraph{Specific models shortcomings.}
We observed an unusual response pattern from GPT-4o when following instructions. Specifically, 79 out of 625 responses (12.64\%) from this model were exactly \textit{[letter]} instead of the expected answer, as dictated by the instructions provided in the prompt:\\

\textit{Instructions:\\
Return the letter corresponding to the correct answer in square brackets.\\
Response format: [letter].\\}
\\
This behaviour was unique to GPT-4o and not evident in other models. Consequently, this led to a decrease in accuracy of 12.6\%, combined with the 5.4\% reduction due to censorship issues discussed later in Section~\ref{sec:censorship}, significantly impacting the model's performance.

We included Minerva-3B in our evaluation, a model with a modest 3 billion parameters trained on 660 billion tokens. Despite not being instruction-tuned~\cite{wei2022finetuned}, the authors evaluated it on several popular benchmarks. Notably, Minerva-3B was pre-trained on a dataset where 50\% of the data was Italian, highlighting its focus on Italian language tasks. However, despite this pre-training strategy, the relatively small size of the model and its training set pose significant challenges. Our tests revealed that Minerva-3B achieved a modest score of 4.9\% on the benchmark, emphasising that while the pre-training dataset composition is important, the overall size of the dataset and the number of parameters are more crucial for handling complex language tasks. 
This reflects the results and findings of other benchmarks. 

\subsection{Limitations}
Various limitations were encountered throughout this research:

\paragraph{Data availability.}
The dataset obtained from Gestinv includes all the INVALSI tests on the Italian language; however, a few questions (3 or 4) were missing from certain tests. Moreover, in some tests, particularly those labelled as simulations, a few questions were missing multiple-choice options, rendering the questions unclear. Some metadata was wrongly labelled. These minor issues were identified and rectified through manual intervention.

\paragraph{OpenAI model censorship.}
\label{sec:censorship}
Some models, particularly OpenAI's GPT models, could not respond to the questions because the prompts were flagged for violence and harassment, preventing them from providing answers. This issue arose due to the context of the questions, which were based on Gianni Rodari's tale "Il padrone della Luna" and Ennio Flaiano's story "Le ombre bianche". The first story depicts a tyrant who uses threats, physical abuse, and oppressive decrees to exploit and torment his subjects. The second tale contains mocking and perpetuating negative stereotypes about a group of people or professions. As a result, evaluating these 34 answers using GPT models was not feasible, leading us to mark them as incorrect, which caused a potential loss of up to 5.4\% in overall accuracy.

\paragraph{Potential shortcomings in complex answer evaluation.}
A subset of questions (seven in total) posed a significant challenge in the evaluation due to the requirement for subjective judgment to determine the correctness of the answers. 
These questions necessitate that the generated answers be semantically relevant to the target answers provided as references. The complexity arises because semantic relevance is not always easily quantifiable, leading to potential inconsistencies in assessment.
To address this, we employed BERTscore~\cite{zhang2019bertscore} to establish an empirical threshold where answers with a BERT score greater than 0.70 were considered correct, while those below this threshold were deemed incorrect. While this method provided a systematic evaluation approach, it has limitations. 
In practice, this method has been manually validated to work well in all present cases, and future cases will be carefully monitored.

\section{Conclusion and Future Work}
\label{sec:conclusion}

This research paper proposes a new benchmark for evaluating proficiency in large language models by structuring the Italian INVALSI tests. We offer several key contributions. First, we establish a structured benchmark specifically for the Italian language, which is a considerable step given the dominance of English-based evaluations in the field. Second, an extensive assessment of current LLMs is performed, offering valuable insights into the capabilities and limitations of these models in understanding and processing Italian. This evaluation generalises to multiple languages, as it assesses general knowledge and the ability to comprehend and reconstruct text meaning and answer questions requiring inferential logical steps. The models' performances are graphically compared across various dimensions and against human performance, highlighting the specific strengths of each model. This comprehensive analysis serves the academic community by providing a robust reference and invites ongoing submissions for model evaluation, ensuring the benchmark remains a dynamic and up-to-date resource.

\paragraph{Main findings.} 
The main findings we have found in our work are:
\begin{itemize}
    \item Models perform better on tasks aimed at lower school grades than those designed for higher grades. The complexity of language and cognitive tasks in higher educational levels poses significant challenges for current language models. Models excel in text comprehension while reflecting on the Italian language is harder.
    \item Larger models consistently outperform smaller ones, even those fine-tuned for the Italian language. This indicates that the inherent capabilities of larger models, possibly due to more extensive training data and more complex neural architectures, contribute to better handling of the nuances of language tasks.
\end{itemize}

Looking ahead, the research aims to expand the scope and utility of the benchmark in several significant ways:

\paragraph{Incorporating mathematics and multimodal capabilities.} The benchmark will be extended to include the mathematical tests, which include multimodal data processing (e.g., geometry problems). This will test the models' abilities to handle linguistic tasks and quantitative and visual information, reflecting more realistic usage scenarios and enhancing their applicability in diverse educational and professional settings.

\paragraph{Increasing the test size.} Additional tests will be incorporated into the benchmark to further the robustness of the evaluations. More tests and questions will lead to less variance in the results, providing a more comprehensive assessment of the LLMs' linguistic capabilities.

\paragraph{Opening up the work to the public with a leaderboard.} The project introduces a public leaderboard to foster a collaborative and competitive research environment. This platform allows researchers worldwide to submit their models for evaluation against the benchmark. Leveraging the global research community's collective expertise will enhance transparency and encourage continuous improvement and innovation in the field. The leaderboard is currently available at \url{https://huggingface.co/spaces/Crisp-Unimib/INVALSIbenchmark}.

These initiatives are poised to significantly enhance the benchmark's relevance and effectiveness, driving forward the evaluation standards for language models in Italian.

\appendix



\section*{Declaration of Interest}
The authors declare that they have no competing interests.

\section*{Ethical Statement}

There are no ethical issues.

\section*{Acknowledgments}

Evaluation of the open-source models was conducted on Leonardo supercomputer with the support of CINECA-Italian Super Computing Resource Allocation, class C project
IsCb7\_LLM-EVAL (HP10CIO7T9).

\bibliographystyle{named}
\bibliography{bibliography}

\begin{thebibliography}{}

\bibitem[\protect\citeauthoryear{Achiam \bgroup \em et al.\egroup }{2023}]{achiam2023gpt}
Josh Achiam, Steven Adler, Sandhini Agarwal, Lama Ahmad, Ilge Akkaya, Florencia~Leoni Aleman, Diogo Almeida, Janko Altenschmidt, Sam Altman, Shyamal Anadkat, et~al.
\newblock Gpt-4 technical report.
\newblock {\em arXiv preprint arXiv:2303.08774}, 2023.

\bibitem[\protect\citeauthoryear{AI@Meta}{2024}]{llama3modelcard}
AI@Meta.
\newblock Llama 3 model card.
\newblock 2024.

\bibitem[\protect\citeauthoryear{Armengol-Estap{\'e} \bgroup \em et al.\egroup }{2021}]{armengol2021multilingual}
Jordi Armengol-Estap{\'e}, Ona de~Gibert Bonet, and Maite Melero.
\newblock On the multilingual capabilities of very large-scale english language models.
\newblock {\em arXiv preprint arXiv:2108.13349}, 2021.

\bibitem[\protect\citeauthoryear{Bacciu \bgroup \em et al.\egroup }{2023}]{bacciu2023fauno}
Andrea Bacciu, Giovanni Trappolini, Andrea Santilli, Emanuele Rodol{\`a}, and Fabrizio Silvestri.
\newblock Fauno: The italian large language model that will leave you senza parole!
\newblock {\em arXiv preprint arXiv:2306.14457}, 2023.

\bibitem[\protect\citeauthoryear{Basile \bgroup \em et al.\egroup }{2023a}]{basile2023llamantino}
Pierpaolo Basile, Elio Musacchio, Marco Polignano, Lucia Siciliani, Giuseppe Fiameni, and Giovanni Semeraro.
\newblock Llamantino: Llama 2 models for effective text generation in italian language.
\newblock {\em arXiv preprint arXiv:2312.09993}, 2023.

\bibitem[\protect\citeauthoryear{Basile \bgroup \em et al.\egroup }{2023b}]{basile2023uinauil}
Valerio Basile, Livio Bioglio, Alessio Bosca, Cristina Bosco, and Viviana Patti.
\newblock Uinauil: A unified benchmark for italian natural language understanding.
\newblock In {\em Proceedings of the 61st Annual Meeting of the Association for Computational Linguistics (Volume 3: System Demonstrations)}, pages 348--356, 2023.

\bibitem[\protect\citeauthoryear{Bolondi \bgroup \em et al.\egroup }{2017}]{bolondi2017database}
Giorgio Bolondi, Alessandro Gambini, and Federica Ferretti.
\newblock Il database gestinv delle prove standardizzate invalsi: Uno strumento per la ricerca: Alcuni esempi di utilizzo nell'ambito della matematica.
\newblock In {\em I dati INVALSI: Uno strumento per la ricerca}, pages 43--48. Franco Angeli, 2017.

\bibitem[\protect\citeauthoryear{Chang \bgroup \em et al.\egroup }{2023}]{chang_survey_2023}
Yupeng Chang, Xu~Wang, Jindong Wang, Yuan Wu, Linyi Yang, Kaijie Zhu, Hao Chen, Xiaoyuan Yi, Cunxiang Wang, Yidong Wang, Wei Ye, Yue Zhang, Yi~Chang, Philip~S. Yu, Qiang Yang, and Xing Xie.
\newblock A survey on evaluation of large language models, 2023.

\bibitem[\protect\citeauthoryear{Chung \bgroup \em et al.\egroup }{2020}]{chung2020improving}
Hyung~Won Chung, Dan Garrette, Kiat~Chuan Tan, and Jason Riesa.
\newblock Improving multilingual models with language-clustered vocabularies.
\newblock {\em arXiv preprint arXiv:2010.12777}, 2020.

\bibitem[\protect\citeauthoryear{Clark \bgroup \em et al.\egroup }{2018}]{clark2018think}
Peter Clark, Isaac Cowhey, Oren Etzioni, Tushar Khot, Ashish Sabharwal, Carissa Schoenick, and Oyvind Tafjord.
\newblock Think you have solved question answering? try arc, the ai2 reasoning challenge.
\newblock {\em arXiv preprint arXiv:1803.05457}, 2018.

\bibitem[\protect\citeauthoryear{Cobbe \bgroup \em et al.\egroup }{2021}]{cobbe2021training}
Karl Cobbe, Vineet Kosaraju, Mohammad Bavarian, Mark Chen, Heewoo Jun, Lukasz Kaiser, Matthias Plappert, Jerry Tworek, Jacob Hilton, Reiichiro Nakano, et~al.
\newblock Training verifiers to solve math word problems, 2021.
\newblock {\em URL https://arxiv. org/abs/2110.14168}, 2021.

\bibitem[\protect\citeauthoryear{Corsini and Losito}{2013}]{corsini2013rilevazioni}
Cristiano Corsini and Bruno Losito.
\newblock Le rilevazioni invalsi: a che cosa servono?
\newblock {\em Cadmo: giornale italiano di pedagogia sperimentale: 2, 2013}, pages 55--76, 2013.

\bibitem[\protect\citeauthoryear{Corsini}{2013}]{corsini2013validita}
Cristiano Corsini.
\newblock La validit{\`a} di contenuto delle prove invalsi di comprensione della lettura.
\newblock {\em Italian Journal of Educational Research}, (10):46--61, 2013.

\bibitem[\protect\citeauthoryear{De~Mattei \bgroup \em et al.\egroup }{2020}]{de2020geppetto}
Lorenzo De~Mattei, Michele Cafagna, Felice Dell'Orletta, Malvina Nissim, and Marco Guerini.
\newblock Geppetto carves italian into a language model.
\newblock {\em arXiv preprint arXiv:2004.14253}, 2020.

\bibitem[\protect\citeauthoryear{Dettmers \bgroup \em et al.\egroup }{2024}]{dettmers2024qlora}
Tim Dettmers, Artidoro Pagnoni, Ari Holtzman, and Luke Zettlemoyer.
\newblock Qlora: Efficient finetuning of quantized llms.
\newblock {\em Advances in Neural Information Processing Systems}, 36, 2024.

\bibitem[\protect\citeauthoryear{Devlin \bgroup \em et al.\egroup }{2019}]{devlin_bert_2019}
Jacob Devlin, Ming-Wei Chang, Kenton Lee, and Kristina Toutanova.
\newblock {BERT}: Pre-training of deep bidirectional transformers for language understanding.
\newblock In {\em Proceedings of the 2019 Conference of the North American Chapter of the Association for Computational Linguistics: Human Language Technologies, Volume 1 (Long and Short Papers)}, pages 4171--4186. Association for Computational Linguistics, 2019.

\bibitem[\protect\citeauthoryear{Guzzo}{2023}]{guzzo2023competenza}
Giulia Guzzo.
\newblock La competenza grammaticale nelle prove invalsi.
\newblock 2023.

\bibitem[\protect\citeauthoryear{Hendrycks \bgroup \em et al.\egroup }{2020}]{hendrycks2020measuring}
Dan Hendrycks, Collin Burns, Steven Basart, Andy Zou, Mantas Mazeika, Dawn Song, and Jacob Steinhardt.
\newblock Measuring massive multitask language understanding.
\newblock {\em arXiv preprint arXiv:2009.03300}, 2020.

\bibitem[\protect\citeauthoryear{Hu \bgroup \em et al.\egroup }{2021}]{hu2021lora}
Edward~J Hu, Yelong Shen, Phillip Wallis, Zeyuan Allen-Zhu, Yuanzhi Li, Shean Wang, Lu~Wang, and Weizhu Chen.
\newblock Lora: Low-rank adaptation of large language models.
\newblock {\em arXiv preprint arXiv:2106.09685}, 2021.

\bibitem[\protect\citeauthoryear{Huang and Chang}{2022}]{huang2022towards}
Jie Huang and Kevin Chen-Chuan Chang.
\newblock Towards reasoning in large language models: A survey.
\newblock {\em arXiv preprint arXiv:2212.10403}, 2022.

\bibitem[\protect\citeauthoryear{Jiang \bgroup \em et al.\egroup }{2023}]{jiang2023mistral}
Albert~Q Jiang, Alexandre Sablayrolles, Arthur Mensch, Chris Bamford, Devendra~Singh Chaplot, Diego de~las Casas, Florian Bressand, Gianna Lengyel, Guillaume Lample, Lucile Saulnier, et~al.
\newblock Mistral 7b.
\newblock {\em arXiv preprint arXiv:2310.06825}, 2023.

\bibitem[\protect\citeauthoryear{Jiang \bgroup \em et al.\egroup }{2024}]{jiang2024mixtral}
Albert~Q Jiang, Alexandre Sablayrolles, Antoine Roux, Arthur Mensch, Blanche Savary, Chris Bamford, Devendra~Singh Chaplot, Diego de~las Casas, Emma~Bou Hanna, Florian Bressand, et~al.
\newblock Mixtral of experts.
\newblock {\em arXiv preprint arXiv:2401.04088}, 2024.

\bibitem[\protect\citeauthoryear{Kaplan \bgroup \em et al.\egroup }{2020}]{kaplan2020scaling}
Jared Kaplan, Sam McCandlish, Tom Henighan, Tom~B Brown, Benjamin Chess, Rewon Child, Scott Gray, Alec Radford, Jeffrey Wu, and Dario Amodei.
\newblock Scaling laws for neural language models.
\newblock {\em arXiv preprint arXiv:2001.08361}, 2020.

\bibitem[\protect\citeauthoryear{Kreutzer \bgroup \em et al.\egroup }{2022}]{kreutzer2022quality}
Julia Kreutzer, Isaac Caswell, Lisa Wang, Ahsan Wahab, Daan van Esch, Nasanbayar Ulzii-Orshikh, Allahsera Tapo, Nishant Subramani, Artem Sokolov, Claytone Sikasote, et~al.
\newblock Quality at a glance: An audit of web-crawled multilingual datasets.
\newblock {\em Transactions of the Association for Computational Linguistics}, 10:50--72, 2022.

\bibitem[\protect\citeauthoryear{Lai \bgroup \em et al.\egroup }{2023}]{lai2023evalita}
Mirko Lai, Stefano Menini, Marco Polignano, Valentina Russo, Rachele Sprugnoli, Giulia Venturi, et~al.
\newblock Evalita 2023: Overview of the 8th evaluation campaign of natural language processing and speech tools for italian.
\newblock In {\em Proceedings of the Eighth Evaluation Campaign of Natural Language Processing and Speech Tools for Italian. Final Workshop (EVALITA 2023), CEUR. org, Parma, Italy}, 2023.

\bibitem[\protect\citeauthoryear{Landro \bgroup \em et al.\egroup }{2022}]{landro2022two}
Nicola Landro, Ignazio Gallo, Riccardo La~Grassa, and Edoardo Federici.
\newblock Two new datasets for italian-language abstractive text summarization.
\newblock {\em Information}, 13(5):228, 2022.

\bibitem[\protect\citeauthoryear{Le~Scao \bgroup \em et al.\egroup }{2023}]{le2023bloom}
Teven Le~Scao, Angela Fan, Christopher Akiki, Ellie Pavlick, Suzana Ili{\'c}, Daniel Hesslow, Roman Castagn{\'e}, Alexandra~Sasha Luccioni, Fran{\c{c}}ois Yvon, Matthias Gall{\'e}, et~al.
\newblock Bloom: A 176b-parameter open-access multilingual language model.
\newblock 2023.

\bibitem[\protect\citeauthoryear{Li \bgroup \em et al.\egroup }{2024}]{li2024eliciting}
Jiahuan Li, Hao Zhou, Shujian Huang, Shanbo Cheng, and Jiajun Chen.
\newblock Eliciting the translation ability of large language models via multilingual finetuning with translation instructions.
\newblock {\em Transactions of the Association for Computational Linguistics}, 12:576--592, 2024.

\bibitem[\protect\citeauthoryear{Liang \bgroup \em et al.\egroup }{2022}]{liang2022holistic}
Percy Liang, Rishi Bommasani, Tony Lee, Dimitris Tsipras, Dilara Soylu, Michihiro Yasunaga, Yian Zhang, Deepak Narayanan, Yuhuai Wu, Ananya Kumar, et~al.
\newblock Holistic evaluation of language models.
\newblock {\em arXiv preprint arXiv:2211.09110}, 2022.

\bibitem[\protect\citeauthoryear{Liang \bgroup \em et al.\egroup }{2023}]{liang2023xlm}
Davis Liang, Hila Gonen, Yuning Mao, Rui Hou, Naman Goyal, Marjan Ghazvininejad, Luke Zettlemoyer, and Madian Khabsa.
\newblock Xlm-v: Overcoming the vocabulary bottleneck in multilingual masked language models.
\newblock {\em arXiv preprint arXiv:2301.10472}, 2023.

\bibitem[\protect\citeauthoryear{Lin \bgroup \em et al.\egroup }{2021}]{lin2021truthfulqa}
Stephanie Lin, Jacob Hilton, and Owain Evans.
\newblock Truthfulqa: Measuring how models mimic human falsehoods.
\newblock {\em arXiv preprint arXiv:2109.07958}, 2021.

\bibitem[\protect\citeauthoryear{Mangrulkar \bgroup \em et al.\egroup }{2022}]{peft}
Sourab Mangrulkar, Sylvain Gugger, Lysandre Debut, Younes Belkada, Sayak Paul, and Benjamin Bossan.
\newblock Peft: State-of-the-art parameter-efficient fine-tuning methods.
\newblock \url{https://github.com/huggingface/peft}, 2022.

\bibitem[\protect\citeauthoryear{Papineni \bgroup \em et al.\egroup }{2002}]{papineni2002bleu}
Kishore Papineni, Salim Roukos, Todd Ward, and Wei-Jing Zhu.
\newblock Bleu: a method for automatic evaluation of machine translation.
\newblock In {\em Proceedings of the 40th annual meeting of the Association for Computational Linguistics}, pages 311--318, 2002.

\bibitem[\protect\citeauthoryear{Pastore \bgroup \em et al.\egroup }{2017}]{pastore2017questione}
Serafina Pastore, Michela Freddano, et~al.
\newblock “questione di feedback”: dati invalsi e pratiche di valutazione in classe.
\newblock In {\em I dati INVALSI: uno strumento per la ricerca}, pages 89--100. FrancoAngeli, 2017.

\bibitem[\protect\citeauthoryear{Pires \bgroup \em et al.\egroup }{2019}]{pires2019multilingual}
Telmo Pires, Eva Schlinger, and Dan Garrette.
\newblock How multilingual is multilingual bert?
\newblock {\em arXiv preprint arXiv:1906.01502}, 2019.

\bibitem[\protect\citeauthoryear{Polignano \bgroup \em et al.\egroup }{2024}]{polignano2024advanced}
Marco Polignano, Pierpaolo Basile, and Giovanni Semeraro.
\newblock Advanced natural-based interaction for the italian language: Llamantino-3-anita, 2024.

\bibitem[\protect\citeauthoryear{Radford \bgroup \em et al.\egroup }{2018}]{radford_improving_2018}
Alec Radford, Karthik Narasimhan, Tim Salimans, and Ilya Sutskever.
\newblock Improving language understanding by generative pre-training.
\newblock 2018.

\bibitem[\protect\citeauthoryear{Rajpurkar \bgroup \em et al.\egroup }{2016}]{rajpurkar2016squad}
Pranav Rajpurkar, Jian Zhang, Konstantin Lopyrev, and Percy Liang.
\newblock Squad: 100,000+ questions for machine comprehension of text.
\newblock {\em arXiv preprint arXiv:1606.05250}, 2016.

\bibitem[\protect\citeauthoryear{Reid \bgroup \em et al.\egroup }{2024}]{reid2024gemini}
Machel Reid, Nikolay Savinov, Denis Teplyashin, Dmitry Lepikhin, Timothy Lillicrap, Jean-baptiste Alayrac, Radu Soricut, Angeliki Lazaridou, Orhan Firat, Julian Schrittwieser, et~al.
\newblock Gemini 1.5: Unlocking multimodal understanding across millions of tokens of context.
\newblock {\em arXiv preprint arXiv:2403.05530}, 2024.

\bibitem[\protect\citeauthoryear{Ruder \bgroup \em et al.\egroup }{2021}]{ruder2021xtreme}
Sebastian Ruder, Noah Constant, Jan Botha, Aditya Siddhant, Orhan Firat, Jinlan Fu, Pengfei Liu, Junjie Hu, Dan Garrette, Graham Neubig, et~al.
\newblock Xtreme-r: Towards more challenging and nuanced multilingual evaluation.
\newblock In {\em Proceedings of the 2021 Conference on Empirical Methods in Natural Language Processing}. Association for Computational Linguistics, 2021.

\bibitem[\protect\citeauthoryear{Sakaguchi \bgroup \em et al.\egroup }{2021}]{sakaguchi2021winogrande}
Keisuke Sakaguchi, Ronan~Le Bras, Chandra Bhagavatula, and Yejin Choi.
\newblock Winogrande: An adversarial winograd schema challenge at scale.
\newblock {\em Communications of the ACM}, 64(9):99--106, 2021.

\bibitem[\protect\citeauthoryear{Santilli and Rodol{\`a}}{2023}]{santilli2023camoscio}
Andrea Santilli and Emanuele Rodol{\`a}.
\newblock Camoscio: An italian instruction-tuned llama.
\newblock {\em arXiv preprint arXiv:2307.16456}, 2023.

\bibitem[\protect\citeauthoryear{Srivastava \bgroup \em et al.\egroup }{2022}]{srivastava2022beyond}
Aarohi Srivastava, Abhinav Rastogi, Abhishek Rao, Abu Awal~Md Shoeb, Abubakar Abid, Adam Fisch, Adam~R Brown, Adam Santoro, Aditya Gupta, Adri{\`a} Garriga-Alonso, et~al.
\newblock Beyond the imitation game: Quantifying and extrapolating the capabilities of language models.
\newblock {\em arXiv preprint arXiv:2206.04615}, 2022.

\bibitem[\protect\citeauthoryear{Talat \bgroup \em et al.\egroup }{2022}]{talat2022you}
Zeerak Talat, Aur{\'e}lie N{\'e}v{\'e}ol, Stella Biderman, Miruna Clinciu, Manan Dey, Shayne Longpre, Sasha Luccioni, Maraim Masoud, Margaret Mitchell, Dragomir Radev, et~al.
\newblock You reap what you sow: On the challenges of bias evaluation under multilingual settings.
\newblock In {\em Proceedings of BigScience Episode\# 5--Workshop on Challenges \& Perspectives in Creating Large Language Models}, pages 26--41, 2022.

\bibitem[\protect\citeauthoryear{Team \bgroup \em et al.\egroup }{2023}]{team2023gemini}
Gemini Team, Rohan Anil, Sebastian Borgeaud, Yonghui Wu, Jean-Baptiste Alayrac, Jiahui Yu, Radu Soricut, Johan Schalkwyk, Andrew~M Dai, Anja Hauth, et~al.
\newblock Gemini: a family of highly capable multimodal models.
\newblock {\em arXiv preprint arXiv:2312.11805}, 2023.

\bibitem[\protect\citeauthoryear{T{\'o}th}{2023}]{toth2023riflettere}
Zuzana T{\'o}th.
\newblock Riflettere sulle parole: la formazione delle parole nelle prove invalsi.
\newblock {\em Lingue antiche e moderne}, 12:277--298, 2023.

\bibitem[\protect\citeauthoryear{Touvron \bgroup \em et al.\egroup }{2023}]{touvron2023llama2}
Hugo Touvron, Louis Martin, Kevin Stone, Peter Albert, Amjad Almahairi, Yasmine Babaei, Nikolay Bashlykov, Soumya Batra, Prajjwal Bhargava, Shruti Bhosale, et~al.
\newblock Llama 2: Open foundation and fine-tuned chat models.
\newblock {\em arXiv preprint arXiv:2307.09288}, 2023.

\bibitem[\protect\citeauthoryear{Trinchero}{2014}]{trinchero2014servizio}
Roberto Trinchero.
\newblock Il servizio nazionale di valutazione e le prove invalsi. stato dell’arte e proposte per una valutazione come agente di cambiamento.
\newblock {\em Form@ re-Open Journal per la formazione in rete}, 14(4):34--49, 2014.

\bibitem[\protect\citeauthoryear{Tunstall \bgroup \em et al.\egroup }{2023}]{tunstall2023zephyr}
Lewis Tunstall, Edward Beeching, Nathan Lambert, Nazneen Rajani, Kashif Rasul, Younes Belkada, Shengyi Huang, Leandro von Werra, Clémentine Fourrier, Nathan Habib, Nathan Sarrazin, Omar Sanseviero, Alexander~M. Rush, and Thomas Wolf.
\newblock Zephyr: Direct distillation of lm alignment, 2023.

\bibitem[\protect\citeauthoryear{Valmeekam \bgroup \em et al.\egroup }{2022}]{valmeekam2022large}
Karthik Valmeekam, Alberto Olmo, Sarath Sreedharan, and Subbarao Kambhampati.
\newblock Large language models still can't plan (a benchmark for llms on planning and reasoning about change).
\newblock {\em arXiv preprint arXiv:2206.10498}, 2022.

\bibitem[\protect\citeauthoryear{Vaswani \bgroup \em et al.\egroup }{2017}]{vaswani_attention_2017}
Ashish Vaswani, Noam Shazeer, Niki Parmar, Jakob Uszkoreit, Llion Jones, Aidan~N Gomez, {\L}ukasz Kaiser, and Illia Polosukhin.
\newblock Attention is all you need.
\newblock {\em Advances in neural information processing systems}, 30, 2017.

\bibitem[\protect\citeauthoryear{Wei \bgroup \em et al.\egroup }{2022}]{wei2022finetuned}
Jason Wei, Maarten~Paul Bosma, Vincent Zhao, Kelvin Guu, Adams~Wei Yu, Brian Lester, Nan Du, Andrew~Mingbo Dai, and Quoc~V. Le.
\newblock Finetuned language models are zero-shot learners.
\newblock 2022.

\bibitem[\protect\citeauthoryear{Zellers \bgroup \em et al.\egroup }{2019}]{zellers2019hellaswag}
Rowan Zellers, Ari Holtzman, Yonatan Bisk, Ali Farhadi, and Yejin Choi.
\newblock Hellaswag: Can a machine really finish your sentence?
\newblock {\em arXiv preprint arXiv:1905.07830}, 2019.

\bibitem[\protect\citeauthoryear{Zhang \bgroup \em et al.\egroup }{2019}]{zhang2019bertscore}
Tianyi Zhang, Varsha Kishore, Felix Wu, Kilian~Q Weinberger, and Yoav Artzi.
\newblock Bertscore: Evaluating text generation with bert.
\newblock {\em arXiv preprint arXiv:1904.09675}, 2019.

\end{thebibliography}

\end{document}